\documentclass[twocolumn,10pt,a4paper]{article}

\title{Classification Uncertainty of Deep Neural Networks Based on Gradient Information}
\date{}

\newcommand{\changed}[1]{\textcolor{black}{#1}}

\usepackage[utf8]{inputenc}
\usepackage[english]{babel}
\usepackage{amsmath,amsfonts,amsthm,mathtools}
\usepackage{csvsimple}
\usepackage{multicol}
\usepackage{authblk}
\usepackage{makecell}

\author[1]{Philipp Oberdiek}
\author[2]{Matthias Rottmann}
\author[2]{Hanno Gottschalk}
\affil[1]{Technische Universität Dortmund, 44227 Dortmund, Germany\authorcr philipp.oberdiek@udo.edu}
\affil[2]{Bergische Universität Wuppertal, 42119 Wuppertal, Germany\authorcr \{rottmann, hanno.gottschalk\}@uni-wuppertal.de}

\usepackage{algorithm}
\usepackage[vlined,boxruled,linesnumbered,algo2e,algosection]{algorithm2e}
\usepackage{graphicx}
\usepackage{subfigure}
\usepackage{placeins}
\usepackage{float}
\usepackage{xcolor}
\usepackage{tikz}
\usepackage[colorlinks=true,
citecolor=black!50!green,
linkcolor=black!33!red,
]{hyperref}
\usepackage[
style=authoryear,
url=false,
doi=false,
isbn=false,
eprint=false,
mincitenames=1,
maxcitenames=1,
uniquelist=false,
backend=biber]{biblatex}
\usepackage{todonotes}
\DeclareCiteCommand{\cite}
  {\usebibmacro{prenote}}
  {\usebibmacro{citeindex}%
   \printtext[bibhyperref]{\usebibmacro{cite}}}
  {\multicitedelim}
  {\usebibmacro{postnote}}

\DeclareCiteCommand*{\cite}
  {\usebibmacro{prenote}}
  {\usebibmacro{citeindex}%
   \printtext[bibhyperref]{\usebibmacro{citeyear}}}
  {\multicitedelim}
  {\usebibmacro{postnote}}

\DeclareCiteCommand{\parencite}[\mkbibparens]
  {\usebibmacro{prenote}}
  {\usebibmacro{citeindex}%
    \printtext[bibhyperref]{\usebibmacro{cite}}}
  {\multicitedelim}
  {\usebibmacro{postnote}}

\DeclareCiteCommand*{\parencite}[\mkbibparens]
  {\usebibmacro{prenote}}
  {\usebibmacro{citeindex}%
    \printtext[bibhyperref]{\usebibmacro{citeyear}}}
  {\multicitedelim}
  {\usebibmacro{postnote}}

\DeclareCiteCommand{\footcite}[\mkbibfootnote]
  {\usebibmacro{prenote}}
  {\usebibmacro{citeindex}%
  \printtext[bibhyperref]{ \usebibmacro{cite}}}
  {\multicitedelim}
  {\usebibmacro{postnote}}

\DeclareCiteCommand{\footcitetext}[\mkbibfootnotetext]
  {\usebibmacro{prenote}}
  {\usebibmacro{citeindex}%
   \printtext[bibhyperref]{\usebibmacro{cite}}}
  {\multicitedelim}
  {\usebibmacro{postnote}}

\DeclareCiteCommand{\textcite}
  {\boolfalse{cbx:parens}}
  {\usebibmacro{citeindex}%
   \printtext[bibhyperref]{\usebibmacro{textcite}}}
  {\ifbool{cbx:parens}
     {\bibcloseparen\global\boolfalse{cbx:parens}}
     {}%
   \multicitedelim}
  {\usebibmacro{textcite:postnote}}

\usepackage{csquotes}
\usepackage[nameinlink]{cleveref}
\usetikzlibrary{intersections,plotmarks}

\usepackage{titlesec}
\titleformat*{\section}{\large\bfseries}
\titlespacing*{\section}{0pt}{0.8\baselineskip}{0.2\baselineskip}
\titlespacing*{\subsection}{0pt}{0.8\baselineskip}{0.2\baselineskip}
\titlespacing*{\paragraph}{0pt}{0.4\baselineskip}{0.4\baselineskip}

\renewbibmacro{in:}{%
  \ifboolexpr{%
     test {\ifentrytype{article}}%
     or
     test {\ifentrytype{inproceedings}}%
  }{}{\printtext{\bibstring{in}\intitlepunct}}%
}

\DeclareFieldFormat{titlecase}{\MakeTitleCase{#1}}

\newrobustcmd{\MakeTitleCase}[1]{%
  \ifthenelse{\ifcurrentfield{booktitle}\OR\ifcurrentfield{booksubtitle}%
    \OR\ifcurrentfield{maintitle}\OR\ifcurrentfield{mainsubtitle}%
    \OR\ifcurrentfield{journaltitle}\OR\ifcurrentfield{journalsubtitle}%
    \OR\ifcurrentfield{issuetitle}\OR\ifcurrentfield{issuesubtitle}%
    \OR\ifentrytype{book}\OR\ifentrytype{mvbook}\OR\ifentrytype{bookinbook}%
    \OR\ifentrytype{booklet}\OR\ifentrytype{suppbook}%
    \OR\ifentrytype{collection}\OR\ifentrytype{mvcollection}%
    \OR\ifentrytype{suppcollection}\OR\ifentrytype{manual}%
    \OR\ifentrytype{periodical}\OR\ifentrytype{suppperiodical}%
    \OR\ifentrytype{proceedings}\OR\ifentrytype{mvproceedings}%
    \OR\ifentrytype{reference}\OR\ifentrytype{mvreference}%
    \OR\ifentrytype{report}\OR\ifentrytype{thesis}}
    {#1}
    {\MakeSentenceCase{#1}}
}

\DeclareFieldFormat[article]{titlecase}{\MakeSentenceCase{#1}}

\theoremstyle{plain}% default

\theoremstyle{definition}

\theoremstyle{remark}

\crefalias{AlgoLine}{line}
\crefalias{algocf}{algorithm}

\makeatletter
\let\cref@old@stepcounter\stepcounter
\def\stepcounter#1{%
  \cref@old@stepcounter{#1}%
  \cref@constructprefix{#1}{\cref@result}%
  \@ifundefined{cref@#1@alias}%
    {\def\@tempa{#1}}%
    {\def\@tempa{\csname cref@#1@alias\endcsname}}%
  \protected@edef\cref@currentlabel{%
    [\@tempa][\arabic{#1}][\cref@result]%
    \csname p@#1\endcsname\csname the#1\endcsname}}
\makeatother

\makeatletter
\renewcommand{\algocf@caption@boxruled}{%
  \hrule
  \hbox to \hsize{%
    \vrule\hskip-0.4pt
    \vbox{   
       \vskip\interspacetitleboxruled%
       \unhbox\algocf@capbox\hfill
       \vskip\interspacetitleboxruled
       }%
     \hskip-0.4pt\vrule%
   }\nointerlineskip%
}%
\makeatother

\usepackage{enumerate}
\usepackage{titling}
\DeclareMathOperator*{\argmax}{arg\,max}

\begin{document}
\maketitle
\section*{Abstract}
We study the quantification of uncertainty of Convolutional Neural Networks (CNNs) based on gradient metrics. Unlike the classical softmax entropy, such metrics gather information from all layers of the CNN. We show for the \changed{EMNIST digits} data set that for several such metrics we achieve the same meta classification accuracy -- i.e. \changed{the task of classifying predictions as correct or incorrect} without knowing the actual label -- as for entropy thresholding.
\changed{We apply meta classification to \changed{unknown} concepts (out-of-distribution samples) -- EMNIST/Omniglot letters, CIFAR10 and noise -- and demonstrate that meta classification rates for unknown concepts can be increased when using entropy together with several gradient based metrics as input quantities for a meta classifier}. Meta classifiers only trained on the uncertainty metrics of known concepts, i.e.\ EMNIST digits, usually do not perform equally well for all unknown concepts. If we however allow the meta classifier to be trained on uncertainty metrics for some \changed{out-of-distribution samples}, meta classification for concepts remote from EMNIST digits \changed{(then termed known unknowns)} can be improved considerably.

\section{Introduction}

In recent years deep learning has outperformed other classes of predictive models in many applications. In some of these, e.g. autonomous driving or diagnostics in medicine, the reliability of a prediction is of highest interest. In classification tasks, the thresholding on the highest softmax probability or thresholding on the entropy of the classification distributions (softmax output) are commonly used metrics to quantify \changed{classification uncertainty of neural networks}, see e.g.~\cite{DBLP:journals/corr/HendrycksG16c}. However, misclassification is oftentimes not detected by these metrics and it is also well known that these metrics can be fooled easily. 
Many works demonstrated how an input can be designed to fool a neural network such that it incorrectly classifies the input with high confidence (termed adversarial examples, see e.g.~\cite{DBLP:journals/corr/SzegedyZSBEGF13,Goodfellow14Adversarial,DBLP:journals/corr/KurakinGB16,DBLP:journals/corr/abs-1712-07107}). This underlines the need for measures of uncertainty.

A basic statistical study of the performance of softmax probability thresholding on several datasets was developed in \cite{DBLP:journals/corr/HendrycksG16c}. This work also assigns proper out-of-distribution candidate datasets to many common datasets. For instance a network trained on MNIST is applied to images of handwritten letters, scaled gray scale images from CIFAR10, and different types of noise. This represents a baseline for comparisons.

Using classical approaches from uncertainty quantification for modeling input uncertainty and/or model uncertainty, the detection rate of misclassifications can be improved. 
Using the baseline in \cite{DBLP:journals/corr/HendrycksG16c}, an approach named ODIN, which is based on input uncertainty, was published in \cite{DBLP:journals/corr/LiangLS17}. This approach shows improved results compared to pure softmax probability thresholding.
Uncertainty in the weights of a neural network can be modeled using Bayesian neural networks. A practically feasible approximation to Bayesian neural networks was introduced in~\cite{Gal:2016:DBA:3045390.3045502}, known as Monte-Carlo dropout, which also improves over classical softmax probability thresholding.

Since the softmax removes one dimension from its input by normalization, some works also perform outlier detection on the softmax input (the penultimate layer) and outperform softmax probability thresholding as well, see \cite{DBLP:journals/corr/BendaleB15}.

\changed{In this work we propose a different approach to measure uncertainty of a neural network based on gradient information.
Technically, we compute the gradient of the negative log-likelihood of a single sample during inference where the class argument in the log-likelihood is the predicted class. We then extract compressed representations of the gradients, e.g., the norm of a gradient for a chosen layer. E.g., a large norm of the gradient is interpreted as a sign that, if the prediction would be true, major re-learning would be necessary for the CNN. We interpret this 're-learning-stress' as uncertainty and study the performance of different gradient metrics used in two meta classification tasks: separating correct and incorrect predictions and detecting in- and out-of-distribution samples.}

\changed{The closest approaches to ours are probably \cite{DBLP:journals/corr/HendrycksG16c} and \cite{DBLP:journals/corr/BendaleB15} as they also establish a self evaluation procedure for neural networks. However they only incorporate (non-gradient) metrics for particular layers close to the networks output while we consider gradient metrics extracted from all the layers. Just as \cite{DBLP:journals/corr/HendrycksG16c} and \cite{DBLP:journals/corr/BendaleB15} our approach does not make use of input or model uncertainty. However these approaches, as well as our approach, are somewhat orthogonal to classical uncertainty quantification and should be potentially combinable with input uncertainty and model uncertainty, as used in \cite{DBLP:journals/corr/LiangLS17} and \cite{Gal:2016:DBA:3045390.3045502}, respectively.}

The remainder of this work is structured as follows: 
\changed{First, in \cref{sec:metrics} we introduce (gradient) metrics, the concept of meta classification and threshold independent performance measures for meta classification, AUROC and AUPR, that are used in the experiments. In \cref{sec:theory} we introduce the network architecture and the experiment setup containing the choice of data sets.} We use EMNIST~(\cite{emnist}) digits as a known concept on which the CNN is trained and EMNIST letters, CIFAR10 images as well as different types of noise as unknown/unlearned concepts. Then we statistically investigate the separation performance of our metrics for correct vs.\ incorrect classifications provided by CNNs. 
This is followed by a performance study for the detection of in- and out-of-distribution samples (detection of unlearned concepts) in \cref{sec:unlearnedconcepts}. Therefore we also combine available metrics for training and comparing different meta classifiers. In this section meta classifiers are trained only using known concepts, i.e., EMNIST digits. Afterwards, in \cref{sec:knownunknowns}, we insert unlearned concepts (which therefore become known unknowns) into the training of the meta classifiers. While the softmax baseline achieves an AUROC value of $95.83\%$ our approach gains $0.81\%$ in terms of AUROC and even more in terms of AUPR values.

\section{Entropy, Softmax Baseline and Gradient Metrics}\label{sec:metrics}

Given an input $x \in \mathbb{R}^n$, weights $w \in \mathbb{R}^p$ and class labels $y \in \mathcal{C} = \{1,\ldots,q\} $, we denote the output of a neural network
by $f(y|x,w) \in [0,1]$.
The entropy \changed{of the estimated class distribution conditioned on the input} (also called Shannon information, \cite{shannon1948})
\begin{equation}
E(x,w)=-\frac{1}{\log(q)}\sum_{y\in\mathcal{C}} f(y|x,w)\log(f(y|x,w))\,,
\end{equation}
is a well known dispersion measure and widely used for quantifying classification uncertainty of neural networks. \changed{In the following we will use the term entropy in the sense explained above. Note that this should not be confused with the entropy underlying the (not estimated and joint) statistical distribution of inputs and labels. }
The softmax baseline proposed by \cite{DBLP:journals/corr/HendrycksG16c} is calculated as
\begin{equation}
S(x,w) = \max\limits_{y\in\mathcal{C}} f(y|x,w)\,.
\end{equation}

Using the maximum a posteriori principle (MAP), the predicted class is defined by
\begin{equation}\label{eq:map}
\hat{y}(x,w) := \argmax_{y\in\mathcal{C}}f(y|x,w) \,
\end{equation}
according to the Bayes decision rule~\cite{Berger1980}, or as one hot encoded label $\hat{g}(x,w) \in \{0,1\}^q$ with
\begin{align}\label{eq:onehot}
\hat{g}_k(x,w)&=\begin{dcases}
1, &\hat{y}(x,w)=k\\
0, &\text{else}
\end{dcases}
\end{align}
for $k=1,\ldots,q$.
Given an input sample $x^i$ \changed{with one hot label $y^i$}, predicted class label $\hat{g}^i$ (from \cref{eq:onehot}) and a loss function $L=L(f(y|x^i,w),y^i)$, we can calculate the gradient of the loss function with respect to the weights $\nabla_w L = \nabla_w L(f(y|x^i,w),\hat{g}^i)$. \changed{In our experiments we use the gradient of the negative log-likelihood at the predicted class label, which means 
\begin{align}
\begin{split}
L&=L(f(y|x^i,w),\hat{g}^i)\\
&=-\sum_{y\in\mathcal{C}}\hat{g}^i_y\log\left(f(y|x^i,w)\right)\\
&=-\log\left(f(\hat{y}|x^i,w)\right) \, .\\
\end{split}
\end{align}}
We apply the following metrics to this gradient:
\begin{itemize}
\item Absolute norm ($\|\nabla_{w}L\|_1$)
\item Euclidean norm ($\|\nabla_{w}L\|_2$)
\item Minimum ($\min\left(\nabla_{w}L\right)$)
\item Maximum ($\max\left(\nabla_{w}L\right)$)
\item Mean ($\text{mean}\left(\nabla_{w}L\right)$)
\item Skewness ($\text{skew}\left(\nabla_{w}L\right)$)
\item Kurtosis ($\text{kurt}\left(\nabla_{w}L\right)$)
\end{itemize}
These metrics can \changed{either} be applied layerwise by restricting the gradient to those weights belonging to a single layer in the neural network \changed{or to the whole gradient on all layers}.

\changed{The metrics can be sampled over the input $X$ and conditioned to the event of either correct or incorrect classification. Let $T(w)$ and $F(w)$ denote the subset of correctly and incorrectly classified samples for the network $f(y|x,w)$, respectively.
Given a metric $M$ (e.g.\ the entropy $E$ or any gradient based one), the two conditioned distributions $M(X,w)|_{T(w)}$ and $M(X,w)|_{F(w)}$ are further investigated. For a threshold $t$, we measure $P(M(X,w)< t \, | \, {T(w)})$ and $P(M(X,w)\geq t \, | \, {F(w)}  )$ by sampling $X$. If both probabilities are high, $t$ gives a good separation between correctly and incorrectly classified samples. This concept can be transfered to the detection of out-of-distribution samples by defining these as incorrectly classified. We term this procedure (classifying $M(X,w)< t$ vs.\ $M(X,w)\geq t$) \emph{meta classification}.}

\changed{Since there are many possible ways to compute thresholds $t$, we compute our results threshold independent by using \textbf{A}rea \textbf{U}nder the \textbf{R}eceiver \textbf{O}perating \textbf{C}urve (AUROC) and \textbf{A}rea \textbf{U}nder the \textbf{P}recision \textbf{R}ecall curve (AUPR). For any chosen threshold $t$ we define}

\begin{align*}
TP &= \#\{\text{correctly predicted positive cases}\}\,,\\
TN &= \#\{\text{correctly predicted negative cases}\}\,,\\
FP &= \#\{\text{incorrectly predicted positive cases}\}\,,\\
FN &= \#\{\text{incorrectly predicted negative cases}\}\,.\\
\end{align*}
\changed{and} can compute the quantities
\begin{align*}
&R = TPR = \frac{TP}{TP+FN}\\
&\text{(True positive rate or Recall)}\,,\\
&FPR = \frac{FP}{FP+TN}\\
&\text{(False positive rate)}\,,\\
&P = \frac{TP}{TP+FP}\\
&\text{(Precision)}\,.\\
\end{align*}
When dealing with threshold dependent classification techniques, one calculates $TPR$ ($R$), $FPR$ and $P$ for many different thresholds in the value range of the variable. The AUROC is the area under the receiver operating curve, which has the FPR as ordinate and the TPR as abscissa. The AUPR is the area under the precision recall curve, which has the recall as the ordinate and the precision as abscissa. For more information on these \changed{performance measures} see \cite{DBLP:conf/icml/DavisG06}.\\
The AUPR is in general more informative for datasets with a strong imbalance in positive and negative cases and is sensitive to which class is defined as the positive case. Because of that we are computing the AUPR-In and AUPR-Out, for which the definition of a positive case is reversed. In addition the values of one variable are multiplied by $-1$ to switch between AUPR-In and AUPR-Out as in \cite{DBLP:journals/corr/HendrycksG16c}.

\section{Meta Classification -- a Benchmark between Softmax Probability and Gradient Metrics}\label{sec:theory}

\begin{figure*}[t]
\includegraphics[width=\textwidth]{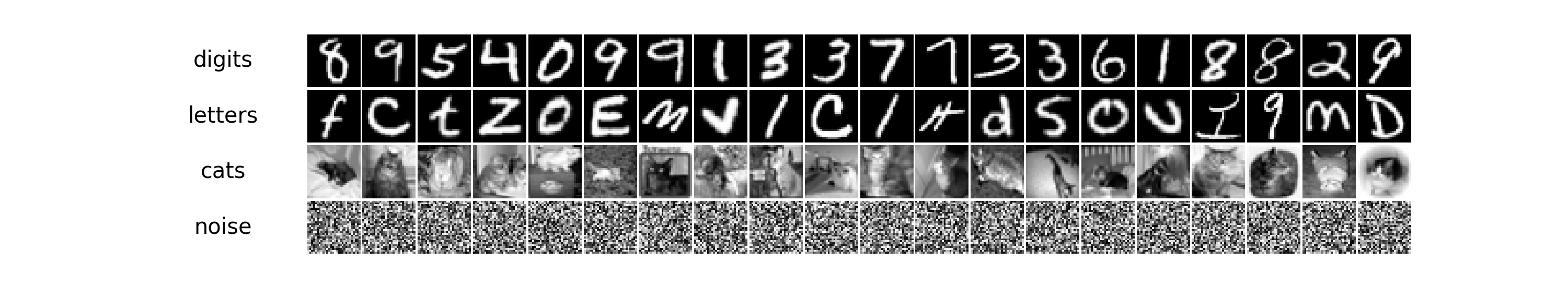}
\caption{Different concepts used for our statistical experiments}
\label{fig:concepts}
\end{figure*}

We perform all our statistical experiments on the EMNIST data set~\cite{emnist}, which contains $28\times 28$ gray scale images of 280\,000 handwritten digits (0 -- 9) and 411\,302 handwritten letters (a -- z, A -- Z). We train the CNNs only on the digits, in order to test their behavior on untrained concepts. We split the EMNIST data set (after a random permutation) as follows:
\begin{itemize}
\item $60,\!000$ digits (0 -- 9) for training
\item $20,\!000$ digits (0 -- 9) for validation
\item $200,\!000$ digits (0 -- 9) for testing
\item $20,\!000$ letters (a -- z, A -- Z) as untrained concepts
\end{itemize}
Additionally we included the CIFAR10 library~\cite{cifar10}, shrinked and converted to gray scale, as well as $20,\!000$ images generated from random uniform noise. All concepts can be seen in \cref{fig:concepts}.

The architecture of the CNNs consists of three convolutional (conv) layers with 16 filters of size $3\times 3$ each, with a stride of 1, as well as a dense layer with a $10$-way softmax output. \changed{Each of the} first two conv layers \changed{are equipped with leaky ReLU activations}
\begin{equation}
LeakyReLU(x)=\begin{dcases}
x, & x>0\\
0.1x, &x<0
\end{dcases}\,
\end{equation}
and followed by $2\times 2$ max pooling. We employ $L^2$ regularization with a regularization parameter of $10^{-3}$. Additionally, dropout~\cite{JMLR:v15:srivastava14a} is applied after the first \changed{and third conv layer}. The dropout rate is $33\%$.

The models are trained using stochastic gradient descent with a batch size of $256$, momentum of $0.9$ and categorical cross entropy as cost function. The initial learning rate is $0.1$ and is reduced by a factor of $10$ every time the average validation accuracy stagnates, until a lower limit for the learning rate of $0.001$ is reached. All models were trained and evaluated using Keras~\cite{chollet2015keras} with Tensorflow backend~\cite{tensorflow2015-whitepaper}. \changed{Note, that the parameters where chosen from experience and not tuned to any extent. The goal is not to achieve a high accuracy, but to detect the uncertainty of a neural network reliably.}

\begin{figure*}[t]
\centering
\includegraphics[width=0.32\textwidth]{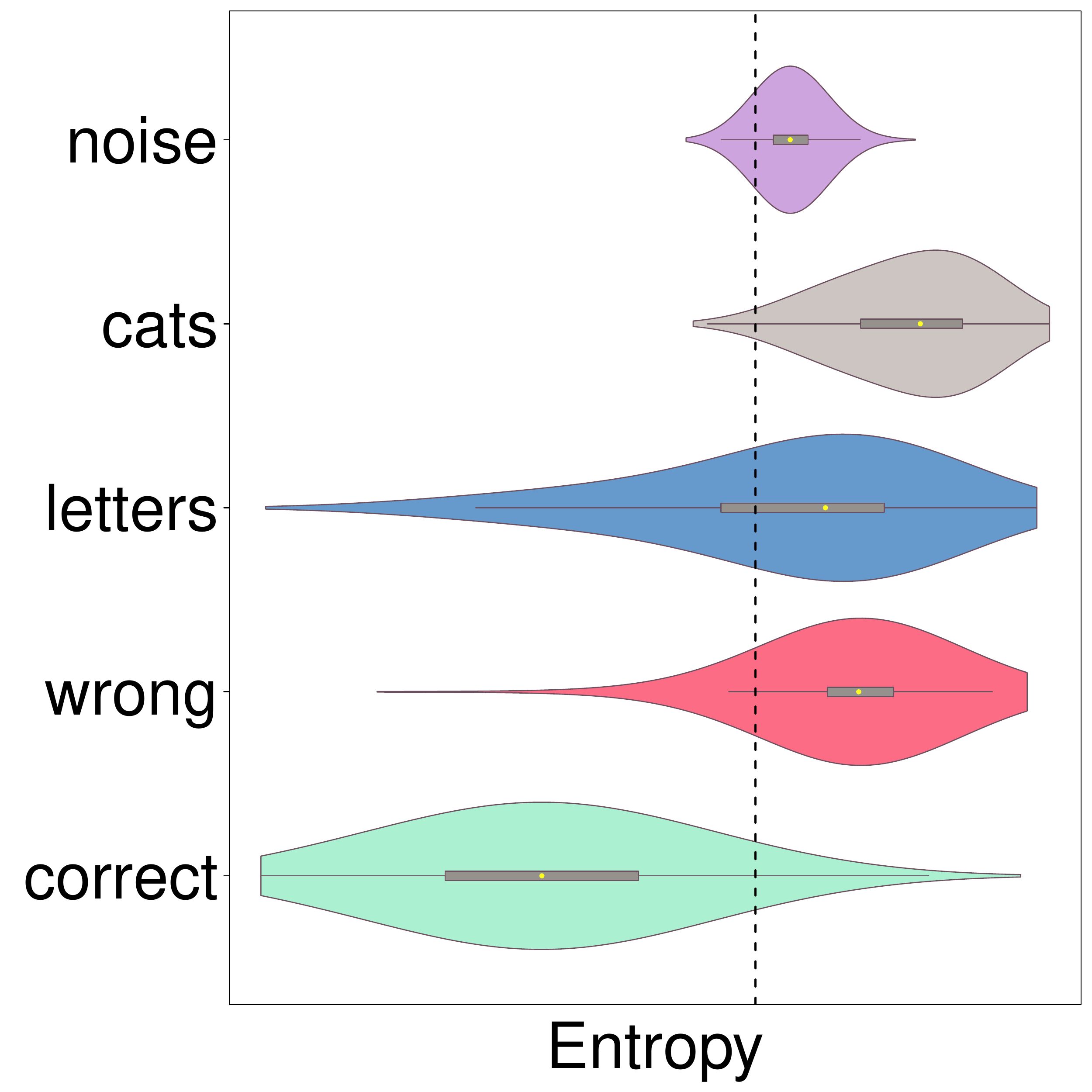}
\includegraphics[width=0.32\textwidth]{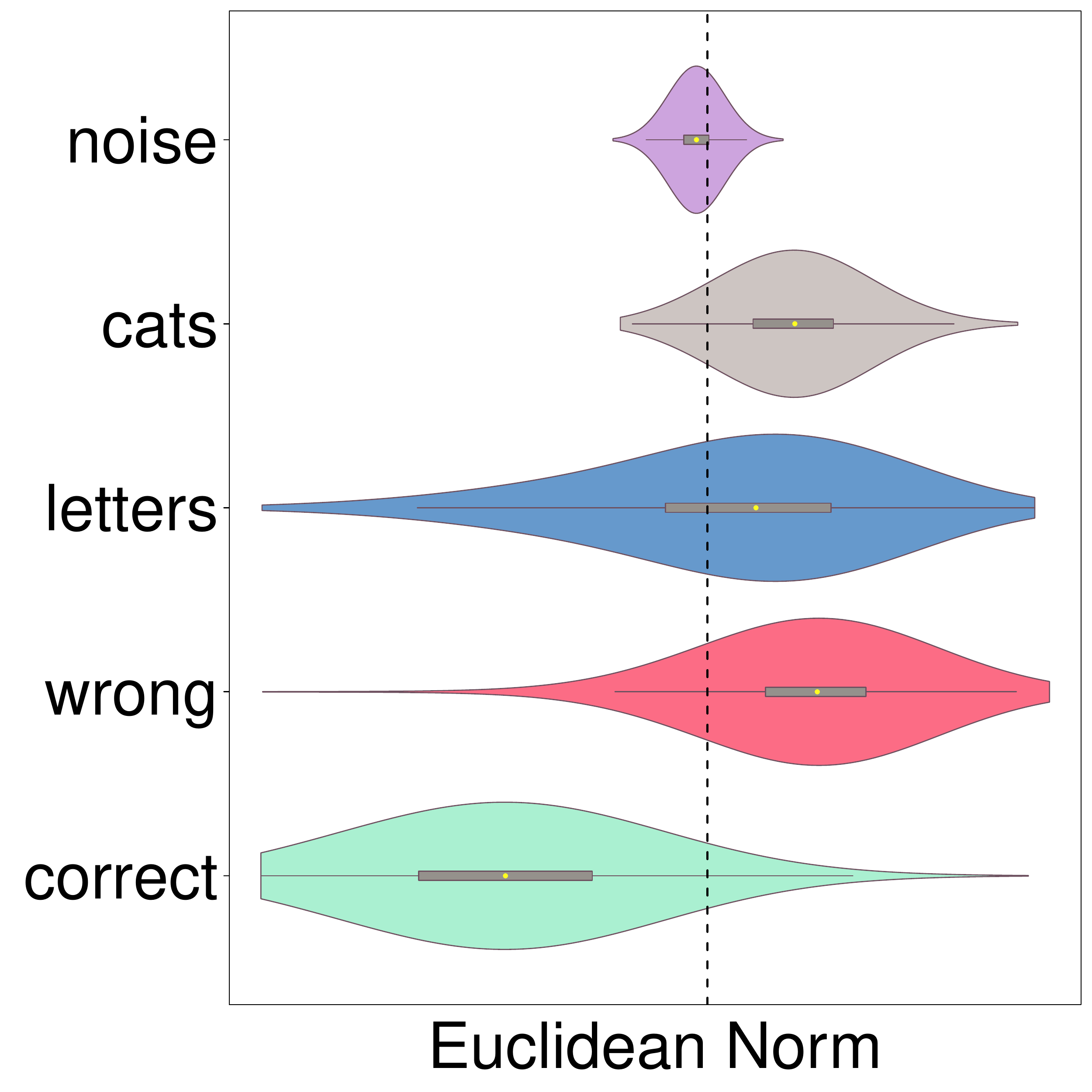}
\includegraphics[width=0.32\textwidth]{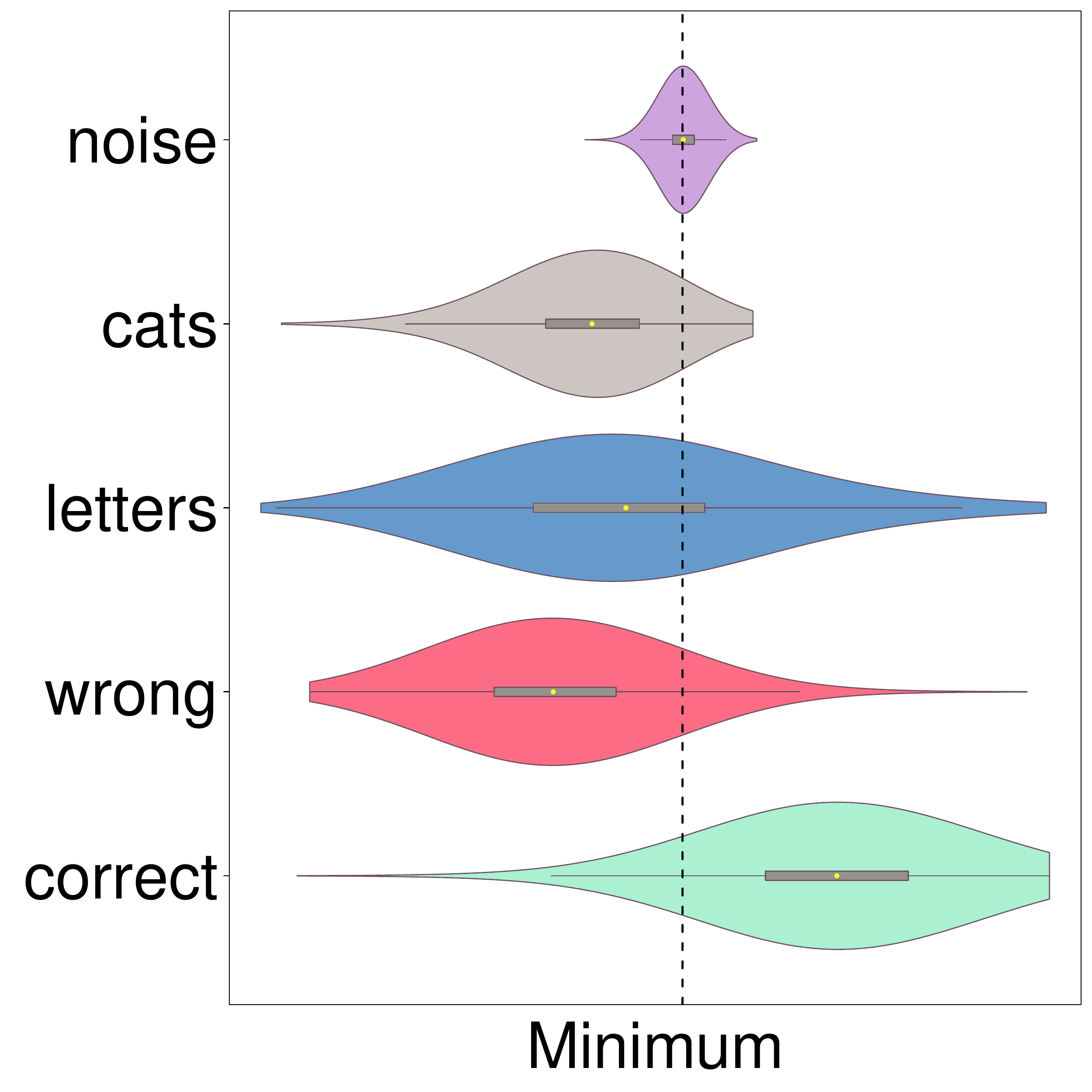}
\caption{Empirical distribution for entropy, euclidean norm and minimum applied to correctly predicted and incorrectly predicted digits from the test data (green and red) of one CNN. Further distributions are generated from EMNIST samples with unlearned letters (blue), CIFAR10 images (gray) and uniform noise images (purple).}
\label{fig:vioall}
\end{figure*}

\changed{In this section, we study the performance of gradient metrics, the softmax baseline and the entropy in terms of AUROC and AUPR for EMNIST test data, thus considering the \emph{error and success prediction} problem, formulated in \cite{DBLP:journals/corr/HendrycksG16c}. First of all we demonstrate that gradient metrics are indeed able to provide good separations. Results for the entropy, euclidean norm and minimum are shown in \cref{fig:vioall} (green and red).
Note that we have left out the mean, skewness and kurtosis metric, as their violin plots showed, that they are not suitable for a threshold meta classifier.}

\changed{In what follows we define EMNISTc as the set containing all correctly classified samples of the EMNIST test set and EMNISTw as the set containing all incorrectly classified ones. From now on we resample the data splitting and use ensembles of CNNs. More precisely, the} random splitting of the $280,\!000$ digit images in training, validation and test data is repeated $10$ times \changed{and we train one CNN for each splitting}. In this way we train $10$ CNNs that differ with respect to initial weights, training, validation and test data. We then repeat the above meta classification for each of the CNNs. With this non parametric bootstrap, we try to get as close as possible to a true sampling of the statistical law underlying the EMNIST ensemble of data and obtain results with statistic validity. 
\begin{table*}[t]
\centering
\begin{tabular}{lcccc}
\hline \thead{Metric} & \thead{EMNISTc /\\ EMNISTw} & \thead{EMNISTc /\\ EMNIST\\ letters} & \thead{EMNISTc /\\ CIFAR10} & \thead{EMNISTc /\\ uniform\\ noise}\\
\hline&\multicolumn{4}{c}{\textbf{AUROC}}\\\hline
\begin{filecontents*}{csv/metrics_results.csv}
rownames,auroccw,auprincw,auproutcw,aurocletters,auprinletters,auproutletters,auroccats,auprincats,auproutcats,aurocnoise,auprinnoise,auproutnoise
Softmax Baseline,\textbf{97.82},\textbf{99.97},39.96,87.62,98.39,59.04,99.13,99.98,77.10,92.95,99.31,40.10
Entropy,97.74,\textbf{99.97},\textbf{95.56},\textbf{88.44},98.38,\textbf{60.36},\textbf{99.42},\textbf{99.99},\textbf{86.07},\textbf{93.52},\textbf{99.36},\textbf{42.46}
Absolute Norm,97.77,\textbf{99.97},95.28,87.22,\textbf{98.42},58.39,98.07,99.95,41.70,90.66,99.07,33.08
Euclidean Norm,97.78,\textbf{99.97},95.30,87.27,\textbf{98.42},58.27,98.34,99.96,47.82,91.05,99.11,34.03
Minimum,97.78,\textbf{99.97},95.36,87.30,95.20,58.76,98.32,99.96,49.32,90.50,99.05,33.00
Maximum,97.70,\textbf{99.97},95.32,86.92,95.03,55.01,98.34,99.96,55.69,87.05,98.67,26.84
Standard Deviation,97.78,\textbf{99.97},95.30,87.26,95.04,58.26,98.34,99.96,47.72,90.98,99.11,33.88
\end{filecontents*}
\csvreader[head to column names, late after line=\\]{csv/metrics_results.csv}{}{%
\rownames & \auroccw & \aurocletters & \auroccats & \aurocnoise}

\hline&\multicolumn{4}{c}{\textbf{AUPR-In}}\\\hline
\csvreader[head to column names, late after line=\\]
{csv/metrics_results.csv}{}{%
\rownames & \auprincw & \auprinletters & \auprincats & \auprinnoise}

\hline&\multicolumn{4}{c}{\textbf{AUPR-Out}}\\\hline
\csvreader[head to column names, late after line=\\]
{csv/metrics_results.csv}{}{%
\rownames & \auproutcw & \auproutletters & \auproutcats & \auproutnoise}
\end{tabular}
\caption{AUROC, AUPR-In (EMNISTc as positive case) and AUPR-Out (EMNISTc as negative case) values for the threshold classification on the softmax baseline, entropy as well as selected gradient metrics. All values are in percentage and averaged over 10 differently initialized CNNs with distinct \changed{splittings} of the training data. \changed{Values in brackets are the standard deviation of the mean in percentage. To get the standard deviation within the sample, multiply by $\sqrt{10}$.}}
\label{tab:aurocauprmetrics}
\end{table*}

\changed{\Cref{tab:aurocauprmetrics} shows that the softmax baseline as well as some selected gradient metrics exhibit comparable performance on the test set in the error and success prediction task. Column one corresponds to the empirical distributions depicted in \cref{fig:vioall} for $200,\!000$ test images.}

In a next step we aggregate entropy and \changed{all} gradient based metrics \changed{(evaluated on the gradient of each layer in the CNN)} in a more sophisticated classification technique. Therefore we choose a variety of regularized and unregularized logistic regression techniques, namely a \textbf{G}eneralized \textbf{L}inear \textbf{M}odel (GLM) equipped with the logit link function, the \textbf{L}east \textbf{A}bsolute \textbf{S}hrinkage and \textbf{S}election \textbf{O}perator (LASSO) with a $L^1$ regularization term and a regularization parameter $\lambda_1=1$, the ridge regression with a $L^2$ regularization term and a regularization parameter of $\lambda_2=1$ and finally the Elastic net with one half $L^1$ and one half $L^2$ regularization, which means $\lambda_1=\lambda_2=0.5$. For details about these methods, cf.~\cite{hastie01statisticallearning}.

To include a non linear classifier we train a feed forward NN with one hidden layer containing $15$ rectified linear units (ReLUs) with $L^2$ weight decay of $10^{-3}$ and $2$-way softmax output. The neural network is trained in the same fashion as the CNNs with stochastic gradient descent. Both groups of classifiers are trained on the EMNIST validation set. Results for the logistic regression techniques can be seen in \cref{tab:regressionresults} (column one) and \changed{those} for the neural network in \cref{tab:nnresults} (first row of each evaluation metric). For comparison we also include the entropy and softmax baseline in each table. The regression techniques perform equally well or better compared to the softmax baseline. This is however not true for the NN. For the logistic regression types including more features from early layers did not improve the performance, the neural network however showed improved results. This means the additional information in those layers can only be utilized by a non linear classifier.

\section{Recognition of Unlearned Concepts}\label{sec:unlearnedconcepts}

A (C)NN, being a statistical classifier, classifies inside the prescribed label space. In this section, we empirically test the hypothesis that test samples out of the label space will be all misclassified, however at a statistically different level of entropy or gradient metric, respectively. We test this hypothesis for three cases: First we feed the CNN with images from the EMNIST letter set and determine the entropy as well as the values for all gradient metrics for each of it. Secondly we follow the same procedure, however the inputs are gray scale CIFAR10 images coarsened to $28\times 28$ pixels. Finally, we use uncorrelated noise that is uniformly distributed in the gray scales with the same resolution. Roughly speaking, we test empirical distributions for unlearned data that is close to the learned concept as in the case of EMNIST letters, data that represents a somewhat remote concept as in the case of CIFAR10 or, as in the case of noise, do not represent any concept at all.

We are classifying the output of a CNN on such input as incorrect label, this way we solve the \emph{in- and out-of-distribution detection problem} from \cite{DBLP:journals/corr/HendrycksG16c}, but are still detecting misclassifications in the prescribed label space.
The empirical distributions of unlearned concepts can be seen in \cref{fig:vioall}. As we can observe, the distributions for incorrectly classified samples are in a statistical sense significantly different from those for correctly classified ones. The gradient metrics however are not able to separate the noise samples very well, but also resulting in an overall good separation of the other concepts, as for the entropy. The threshold classification evaluation metrics can be seen in \cref{tab:aurocauprmetrics}.
\begin{table*}[t]
\centering
\begin{tabular}{lcccc}\hline
\thead{Metric / \\ Regression \\ technique} & \thead{EMNISTc / \\ EMNISTw} & \thead{EMNISTc / \\ EMNIST\\ letters} & \thead{EMNISTc / \\ CIFAR10} & \thead{EMNISTc /\\ uniform\\ noise}\\\hline
&\multicolumn{4}{c}{\textbf{AUROC}}\\\hline
\begin{filecontents*}{csv/logreg_aurocs.csv}
rownames,auroccw,aurocletters,auroccats,aurocnoise
Softmax Baseline,\textbf{97.82} (0.03),87.62 (0.20),99.13 (0.11),92.95 (1.86)
Entropy,97.74 (0.04),88.44 (0.21),\textbf{99.42} (0.10),93.52 (1.83)
GLM,94.76 (0.70),85.94 (0.46),80.26 (5.46),89.41 (2.90)
LASSO,97.75 (0.03),\textbf{89.34} (0.17),99.23 (0.03),93.86 (1.04)
Ridge,97.59 (0.03),88.63 (0.11),98.93 (0.02),\textbf{94.08} (0.67)
Elastic net,97.79 (0.06),89.27 (0.24),98.82 (0.06),93.47 (0.67)
\end{filecontents*}
\csvreader[head to column names, late after line=\\]
{csv/logreg_aurocs.csv}{}
{\rownames & \auroccw & \aurocletters & \auroccats & \aurocnoise}\hline
&\multicolumn{4}{c}{\textbf{AUPR-In}}\\\hline
\begin{filecontents*}{csv/logreg_auprins.csv}
rownames,auprincw,auprinletters,auprincats,auprinnoise
Softmax Baseline,\textbf{99.97} (0.00),\textbf{98.39} (0.03),99.98 (0.00),99.31 (0.19)
Entropy,99.97 (0.00),98.38 (0.04),\textbf{99.99} (0.00),\textbf{99.36} (0.19)
GLM,99.81 (0.05),96.80 (0.21),95.51 (1.15),97.81 (0.84)
LASSO,99.97 (0.00),98.30 (0.06),99.95 (0.00),99.33 (0.12)
Ridge,99.97 (0.00),97.86 (0.04),99.93 (0.00),99.36 (0.08)
Elastic net,99.97 (0.00),98.26 (0.09),99.92 (0.00),99.29 (0.08)
\end{filecontents*}
\csvreader[head to column names, late after line=\\]
{csv/logreg_auprins.csv}{}
{\rownames & \auprincw & \auprinletters & \auprincats & \auprinnoise}\hline
&\multicolumn{4}{c}{\textbf{AUPR-Out}}\\\hline
\begin{filecontents*}{csv/logreg_auprouts.csv}
rownames,auproutcw,auproutletters,auproutcats,auproutnoise
Softmax Baseline,39.96 (0.57),59.04 (0.37),77.10 (1.90),40.10 (4.87)
Entropy,\textbf{95.56} (0.05),60.36 (0.42),86.07 (1.76),42.46 (5.38)
GLM,31.27 (0.79),57.72 (0.74),62.90 (6.77),46.43 (5.24)
LASSO,36.27 (0.32),\textbf{64.04} (0.26),\textbf{91.18} (0.44),\textbf{48.38} (3.12)
Ridge,38.17 (0.34),61.92 (0.18),82.95 (0.61),47.30 (2.20)
Elastic net,38.71 (0.65),63.43 (0.62),79.56 (1.76),45.03 (1.92)
\end{filecontents*}
\csvreader[head to column names, late after line=\\]
{csv/logreg_auprouts.csv}{}
{\rownames & \auproutcw & \auproutletters & \auproutcats & \auproutnoise}
\end{tabular}
\caption{Average AUROC, AUPR-In and AUPR-Out values for different regression types trained on the validation set and all metric features including the entropy \changed{but excluding the softmax baseline}.
The values are averaged over 10 CNNs and displayed in percentage. \changed{The values in brackets are the standard deviations of the mean in percentage. To get the standard deviation within the sample, multiply by $\sqrt{10}$.}}
\label{tab:regressionresults}
\end{table*}
For the logistic regression results in \cref{tab:regressionresults} one can see that the GLM \changed{is inferior to the other methods}. Regression techniques \changed{with a regularization term} like LASSO, Ridge and Elastic net are performing best. We get similar AUROC values as for the threshold classification with single metrics, but can improve between \changed{$5\%$} and \changed{$14.08\%$} over the softmax baseline in terms of AUPR-Out values for unknown concepts, showing a better generalization.

\section{Meta Classification with Known Unknowns}\label{sec:knownunknowns}
In the previous section we trained the meta classifier on the training or validation data only. This means it has no knowledge of entropy or metric distributions for unlearned concepts, hence we followed a puristic approach treating out of distribution cases as unknown unknowns. The classification accuracy could be improved, by \changed{extending} the training set of the meta classifier with the entropy and gradient metric values of a few unlearned concepts \changed{and labeling them as false, i.e., incorrectly predicted. As in the previous sections we then train meta classifiers on the metrics}. For this we use the same data sets as \cite{DBLP:journals/corr/HendrycksG16c}, namely the omniglot handwritten characters set \cite{Omniglot}, the notMNIST dataset \cite{notMNIST} consisting of letters from different fonts, the CIFAR10 dataset \cite{cifar10} coarsened and converted to gray scale as well as normal and uniform noise.
In order to investigate the influence of unknown concepts in the training set of the meta classifier, we used the LASSO regression and the NN introduced in \cref{sec:theory} and supplied them with different training sets, consisting of
\begin{itemize}
\item EMNIST validation set
\item EMNIST validation set and $200$ uniform noise images
\item EMNIST validation set, $200$ uniform noise images and $200$ CIFAR10 images
\item EMNIST validation set, $200$ uniform noise images, $200$ CIFAR10 images and $200$ omniglot images
\end{itemize}
We are omitting the results for the LASSO here, since they are inferior to those of the NN.
\begin{table*}[htb]
\centering
\begin{tabular}{lcccccc}
\hline &&& \multicolumn{4}{c}{Training set for the neural network meta classifier}\\
\thead{Wrong \\ Datasets} & Entropy & \thead{Softmax \\ Baseline \\ \cite{DBLP:journals/corr/HendrycksG16c}} & \thead{EMNIST\\validation} & \thead{EMNIST\\validation+\\uniform\\ noise} & \thead{EMNIST\\validation+\\uniform\\ noise+\\CIFAR10} & \thead{EMNIST\\validation+\\uniform\\ noise+\\CIFAR10+\\omniglot}\\

\hline&\multicolumn{6}{c}{\textbf{AUROC}}\\\hline

\begin{filecontents*}{csv/nn_aurocs.csv}
filler,rownames,aurocsoftmax,aurocval,aurocnoise,aurocnoisecats,aurocnoisecatsomni,aurocentropy
1,EMNISTw,\textbf{97.84},94.59,96.51,96.69,96.68,97.74
2,Omniglot,97.84,94.38,97.29,97.44,97.84,\textbf{98.05}
3,notMNIST,95.24,85.90,93.22,94.49,94.86,\textbf{95.41}
4,CIFAR10,99.03,81.19,96.27,99.12,99.09,\textbf{99.24}
5,Normal noise,94.49,56.09,\textbf{98.37},98.34,98.17,94.36
6,Uniform noise,93.87,86.77,94.22,93.87,\textbf{94.42},94.31
7,All,95.83,80.55,95.49,96.36,\textbf{96.64},96.04
\end{filecontents*}
\csvreader[head to column names, late after line=\\]{csv/nn_aurocs.csv}{}{%
\rownames & \aurocentropy & \aurocsoftmax & \aurocval & \aurocnoise & \aurocnoisecats & \aurocnoisecatsomni}

\hline&\multicolumn{6}{c}{\textbf{AUPR-In}}\\\hline
\begin{filecontents*}{csv/nn_auprins.csv}
filler,rownames,auprinsoftmax,auprinval,auprinnoise,auprinnoisecats,auprinnoisecatsomni,auprinentropy
1,EMNISTw,\textbf{99.97},99.89,99.95,99.96,99.96,99.97
2,Omniglot,99.82,99.04,99.73,99.75,99.80,\textbf{99.84}
3,notMNIST,99.43,95.86,98.83,99.19,99.29,\textbf{99.45}
4,CIFAR10,99.94,95.47,99.41,99.94,99.93,\textbf{99.95}
5,Normal noise,99.60,92.72,\textbf{99.89},99.89,99.88,99.59
6,Uniform noise,99.62,98.05,99.56,99.53,\textbf{99.57},99.65
7,All,98.59,84.98,97.72,98.53,\textbf{98.71},98.66
\end{filecontents*}
\csvreader[head to column names, late after line=\\]{csv/nn_auprins.csv}{}{%
\rownames & \auprinentropy & \auprinsoftmax & \auprinval & \auprinnoise & \auprinnoisecats & \auprinnoisecatsomni}

\hline&\multicolumn{6}{c}{\textbf{AUPR-Out}}\\\hline
\begin{filecontents*}{csv/nn_auprouts.csv}
filler,rownames,auproutsoftmax,auproutval,auproutnoise,auproutnoisecats,auproutnoisecatsomni,auproutentropy
1,EMNISTw,\textbf{39.94},32.95,36.02,35.98,35.33,35.83
2,Omniglot,80.45,74.17,80.36,81.46,83.40,\textbf{83.48}
3,notMNIST,14.59,64.57,71.53,74.91,\textbf{75.13},74.86
4,CIFAR10,87.38,54.45,73.93,90.84,89.82,\textbf{91.27}
5,Normal noise,57.32,18.57,\textbf{68.73},67.89,65.12,54.98
6,Uniform noise,36.63,56.66,58.56,56.59,\textbf{59.53},37.97
7,All,88.07,75.64,89.38,91.23,\textbf{91.55},89.17
\end{filecontents*}
\csvreader[head to column names, late after line=\\]{csv/nn_auprouts.csv}{}{%
\rownames & \auproutentropy & \auproutsoftmax & \auproutval & \auproutnoise & \auproutnoisecats & \auproutnoisecatsomni}
\end{tabular}
\caption{AUROC, AUPR-In (EMNISTc is positive case) and AUPR-Out (EMNISTc is negative case) values for a NN meta classifier. ``All" contains omniglot, notMNIST, CIFAR10, normal noise and uniform noise. We used 200 samples of each concept that was additionally included into the training set. \changed{The supplied features are all gradient based metrics as well as the entropy}. The displayed values are averages over 5 differently initialized NN meta classifiers for each of the 10 CNNs trained on the EMNIST dataset. All values are in percentage \changed{and the values in brackets are the standard deviations of the mean in percentage. To get the mean within the sample multiply by $\sqrt{10}$.}}
\label{tab:nnresults}
\end{table*}
Including known unknowns into the training set, the NN has far better performance on the unknown concepts, even though the amount of additional training data is small.
Noteworthily the validation set together with only 200 uniform noise images increases the results on the AUPR-Out values for all unknown concepts already significantly by $13.74\%$, even comparable to using all concepts. Together with the fact, that noise is virtually available at no cost, it is a very promising candidate for improving the generalization of the meta classifier without the need of generating labels for more datasets. The in-distribution detection rate of correct and wrong predictions is also increased when using additional training concepts, making it only beneficial to include noise into the training set of the meta classifier. Our experiments show however that normal noise does not have such a high influence on the performance as uniform noise and is even decreasing the in-distribution meta classification performance. All in all we reach a $3.48\%$ higher performance on the out of distribution examples compared to the softmax baseline in AUPR-Out and $0.81\%$ in AUROC, whereas the increase in AUPR-In is marginal ($0.12\%$).

 \section{Conclusion and Outlook}
We introduced a new set of metrics that measures the uncertainty of deep CNNs. These metrics have a comparable performance with the widely used entropy and maximum softmax probability to meta-classify whether a certain classification proposed by the underlying CNN is presumably correct or incorrect. Here the performance is measured by AUROC, AUPR-In and AUPR-Out. Entropy and softmax probability perform equally well or slightly better than any single member of the new gradient based metrics for the detection of unknown concepts like EMNIST letters, gray scale converted CIFAR10 images and uniform noise where simple thresholding criteria are applied. \changed{But still, our new metrics allow contributions of different layers and weights to the total uncertainty. } Combining the gradient metrics together with entropy in a more complex meta classifier increases the ability to identify out-of-distribution examples, so that in some cases these meta classifiers outperform the baseline. Additional calibration by including a few samples of unknown concepts increases the performance significantly. Uniform noise proved to raise the overall performance, without the need of more labels. Overall the results for the classification of correct or incorrect predictions increased when the meta classifier was supplied with more distinct concepts in the training set. It seems that the higher number of uncertainty metrics helps to better hedge the correctly classified samples from the variety of out of sample classes, which would be difficult, if only one metric is available. Note that this increase in meta classification is particularly valuable, if one does not want to deteriorate the classification performance of the underlying classifier by additional classes for the known unknowns.  
 
As future work we want to evaluate the performance and robustness of such gradient metrics on different tasks in pattern recognition. Further features could be generated by applying the metrics to activations rather than gradients. One could also investigate the possibility of generating artificial samples, labeled as incorrect, for the training set of the meta classifier in order to further improve the results.
\paragraph{Acknowledgement.}
We thank Fabian H\"uger and Peter Schlicht from Volkswagen Group Research for discussion and remarks on this work. \changed{We also thank the referees for their comments and criticism helping us to improve the paper.}

\newpage
\nocite{*}
\printbibliography
\end{document}